\documentclass[conference]{IEEEtran}

\usepackage{amsmath}
\usepackage{amssymb}
\usepackage{multicol}
\usepackage{cuted}
\usepackage{url}
\usepackage{color}

\usepackage[pdftex]{graphicx,epsfig}

\ifCLASSINFOpdf
\else
\fi
\begin{document}


%
\title{Cascaded Segmentation-Detection Networks for Word-Level Text Spotting}

\author{\IEEEauthorblockN{Siyang Qin, Roberto Manduchi}
\IEEEauthorblockA{Computer Engineering Department, Jack Baskin School of Engineering\\
University of California Santa Cruz\\
Santa Cruz, California 95064\\
Email: siqin@soe.ucsc.edu, manduchi@soe.ucsc.edu }}


%


\maketitle

\begin{abstract}
We introduce an algorithm for word-level text spotting that is able to accurately and reliably determine the bounding regions of individual words of text ``in the wild". Our system is formed by the cascade of two convolutional neural networks. The first network is fully convolutional and is in charge of detecting areas containing text. This results in a very reliable but possibly inaccurate segmentation of the input image. The second network (inspired by the popular YOLO architecture) analyzes each   segment produced in the first stage, and predicts oriented rectangular regions containing individual words. No post-processing (e.g. text line grouping) is necessary. With  execution time of 450 ms for a 1000$\bf \times$560 image on a Titan X GPU, our system achieves the highest score to date among published algorithms on the ICDAR 2015 Incidental Scene Text dataset benchmark \cite{icdarincidental}. 
\end{abstract}


%
\IEEEpeerreviewmaketitle

\section{Introduction}
Fast automatic detection and reading of text (such a license plate number, a posted sign, or a street name) in images taken by a fixed or a moving camera, is very desirable for applications such as surveillance, forensics, autonomous vehicles, augmented reality (e.g., visual translation), and information access for blind people. Traditionally, OCR systems were designed for documents scanned into well-framed, good resolution images without excessive clutter, and taken under good illumination. Recent mobile OCR software implemented  in smartphones (e.g. ABBYY TestGrabber or KNFBReader), dedicated hardware (OrCam), or in the cloud (Google Vision API) produces very good results, but none of these systems is  designed for real-time  deployment (multiple frames per second), which is critical for the applications mentioned above. For computational efficiency, a two-step process is often implemented. The first stage (text spotting) quickly detects the presence of areas in the image that are likely to contain text. These are then passed on to a recognition engine that decodes the textual content, using machine learning normally coupled with lexicon priors.

\begin{figure}[t]
\centering
\includegraphics[width=1\columnwidth]{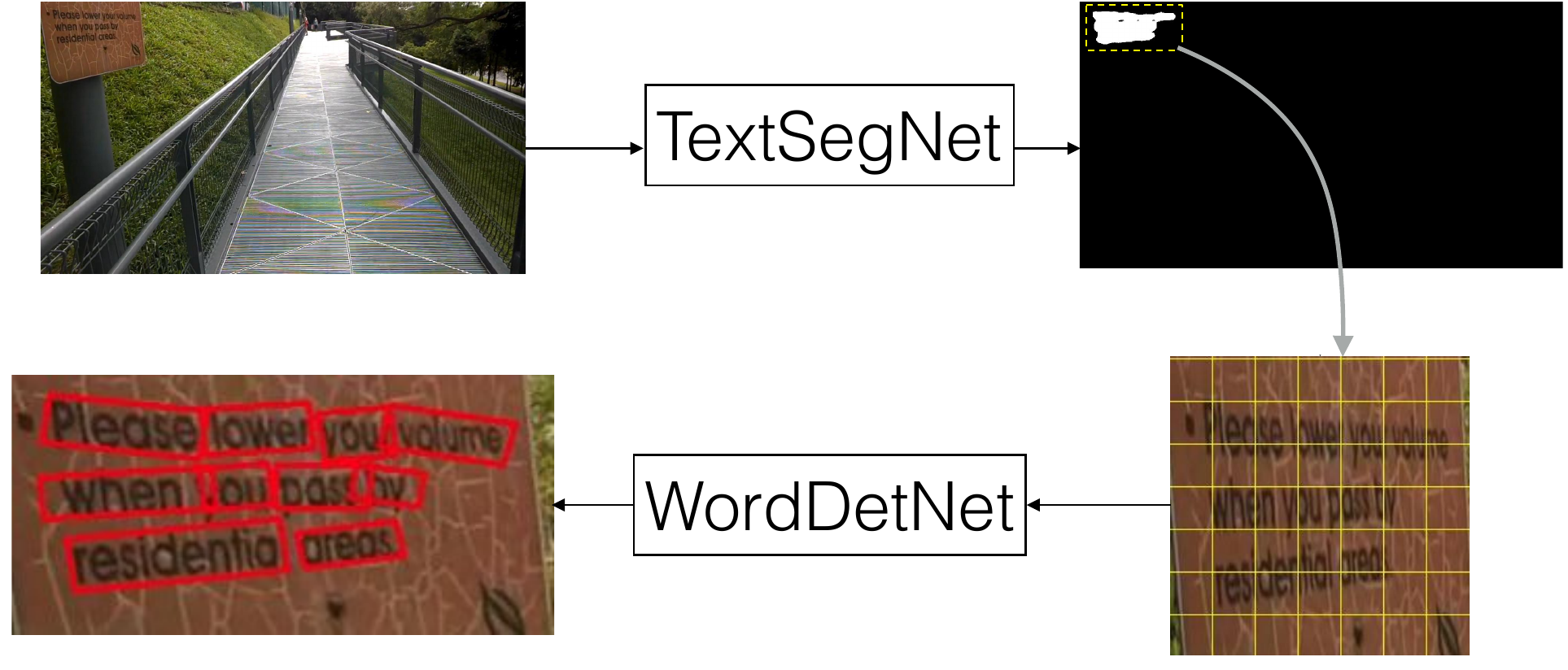}
\caption{The general architecture of our word-level text spotting system. TextSegNet finds text blocks with arbitrary shapes and size. A squared resized block is passed on to WordDetNet, which generates oriented rectangular regions containing individual words.}
\label{fig:fcn-yolo}
\end{figure}

This contribution focuses on fast and accurate word-level text spotting. The ability to detect individual words may simplify the work of the recognizer, and word-level detection is part of typical benchmarks such as the ICDAR incidental and focused datasets \cite{icdarfocused,icdarincidental}. Individual word detection could be cast as an object detection task, for example using popular algorithms such as  as Faster R-CNN \cite{ren2015faster} or YOLO \cite{redmon2016you}, that can directly predict the coordinates of each object using axis-aligned rectangular bounding boxes. Unfortunately, direct application of these algorithms to general text-bearing  images produces unsatisfactory results ~\cite{gupta2016synthetic,tian2016detecting}. This is because general object detection methods have difficulties at detecting groups of very small objects such as words in a text line. 

Another possible approach to text spotting is the use of segmentation algorithms (such as the fully convolutional networks, or FCN~\cite{long2015fully}) to identify images areas that are likely to contain text. These algorithms have proved  very effective in terms of detecting text at variable size, but are generally poor at identifying individual words~\cite{zhang2016multi,yao2016scene}. 

In this work, we combine FCN's remarkable robustness at segmenting text regions, with YOLO's efficient mechanism for detecting objects (in this case, words), appropriately  modeled as oriented rectangles. The two systems are integrated as a cascade (see Fig. \ref{fig:fcn-yolo}):  text regions produced by our fully convolutional network (TextSegNet) are cropped out of the image and resized to a  square shape with fixed size. Then, a YOLO-like network (WordDetNet) is trained to generate oriented rectangular bounding boxes around each word. A simple non-maximum suppression stage takes care of overlapping boxes. In analogy with foveated vision, TextSegNet takes the role of a ``spotter", determining regions of interest to be analyzed in detail by WordDetNet. The resized text regions  contain a limited density of words,  matching the inherently limited capacity of WordDetNet. 
The scheme is simple and elegant,  and requires none of the post-processing steps (region grouping into straight lines, word splitting) that are typical of prior approaches. With execution time of 450 ms per image, our method achieves excellent results on popular benchmarks.

\section{Prior Work}\label{sec:PW}
Early attempts at text spotting used hand-designed features to capture characteristics of text images, both statistical (bimodal marginal brightness distribution) and morphological  (uniform stroke width, connectivity, consistent width and height, alignment into text rows). Two of the most successful examples were \cite{neumann2012real}, based on MSER segmentation, and \cite{epshtein2010detecting}, based on the stroke width transform. While these techniques worked reasonably well, a substantial increase in detection and localization accuracy was achieved  with the use of convolutional neural networks (CNN).  The first such methods~\cite{wang2012end,huang2014robust,zhang2015symmetry,qin2016wacv}  used specific techniques to extract regions ({\em proposals}) with good likelihood of containing text (or individual characters), which were then passed  on to a CNN classifier that would rule out false detections.  
This strategy, however, was plagued by several drawbacks. Detecting individual characters is difficult in the presence of blur, noise, or poor contrast. Since the operators used for character detection were typically local, text-like background elements were often confused with text characters. In order to ensure good recall rate, many (possibly overlapping) templates must be processed by the CNN, resulting in long computational time.


Recent progress in semantic segmentation and object detection has offered new tools that are well suited to text spotting. Fully convolutional networks (FCN) \cite{long2015fully, zheng2015conditional, chen2016deeplab, xie2015holistically, qin2017icme} and end-to-end object detection architectures such as Faster R-CNN \cite{ren2015faster} and YOLO \cite{redmon2016you} process a full image, and produce pixel-wise labelling or labelled regions containing objects of interest. In particular, through the use of skip layers,  FCN are able to analyze an image  using both large and narrow receptive fields, effectively  encoding both local features and global context. Unfortunately, the segmentation produced by FCN, while highly reliable, cannot in general separate individual text lines or words, and further processing is required (see Fig. \ref{fig:FCN_samples}). 

The use of FCN for text spotting was pioneered by Zhang \textit{et al.} \cite{zhang2016multi}, who trained an FCN model to predict a saliency map; text line hypotheses were formed by combining this saliency map with individual character templates found via MSER. A final character-level FCN was used to remove false detections. 
The work of Yao \textit{et al.} \cite{yao2016scene}  added supervision on the scale and center of characters as well as the linking orientation of nearby characters when training the text block FCN. With additional information produced by FCN, the task of the text line grouping module was much simplified. Individual words were found based on the detection of indents  between characters in a text line.

In order to accurately locate text lines,  Tian \textit{et al.} \cite{tian2016detecting} used a recurrent neural network to connect proposal regions into individual lines.  This algorithm could be trained end-to-end; compared to other bottom-up methods, it required no dedicated post processing to form a text line. Unfortunately, this method only worked for near horizontal text lines, and was unable to identify individual words. Gupta \textit{et al.} \cite{gupta2016synthetic} introduced a method for individual word detection that did not require any post-processing (such as grouping individual character templates). They trained a fully convolutional regression network  similar to \cite{redmon2016you} using a large dataset of synthetic images. This algorithm splits an image into cells in a grid, where each cell is responsible for detection of a word  centered in it. Each image cell  predicts the pose parameters (location, width, height, and orientation) of a word, along with a confidence value. This method gives good results on relative simple benchmarks \cite{icdarfocused}, but suffers from its inherent inability  to detect small words in the image, when several such words  locate inside the same (fixed size) image cell. This limits its performance on the challenging ICDAR incidental dataset \cite{icdarincidental}.

For more comprehensive surveys,  the reader is referred to \cite{karatzas2015icdar,karatzas2013icdar,shahab2011icdar,ye2014text}.

\section{Method}
As discussed in Sec.~\ref{sec:PW},  the traditional bottom-up approach (from individual characters to text line grouping to individual words) has recently been  replaced by  top-down strategies, which start by detecting text regions (e.g. using FCN),   then proceed to identifying individual words. Unfortunately, while FCN enables very robust pixel-level text block segmentation, detecting individual text lines or words with the same mechanism becomes more challenging. This can be intuitively justified as follows.

\begin{figure}[h]
\setlength\tabcolsep{1.5pt} 
\begin{tabular}{cc}
\includegraphics[width=0.5\columnwidth]{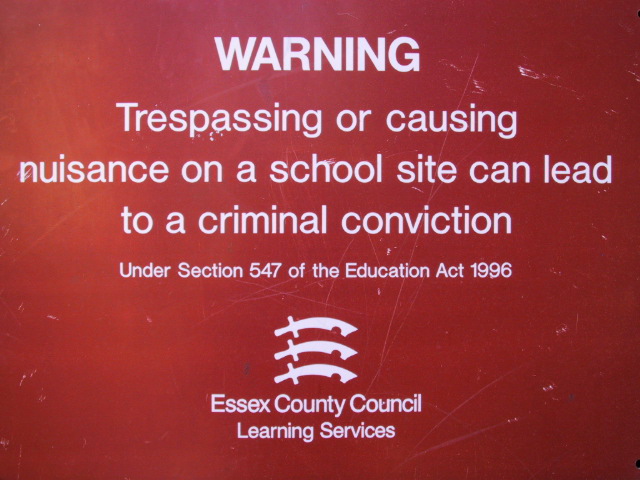}&
\includegraphics[width=0.5\columnwidth]{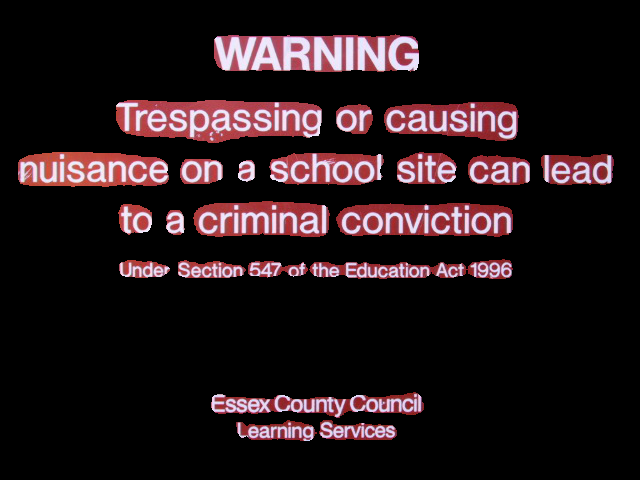}\\
\includegraphics[width=0.5\columnwidth]{figures/img_26.jpg}&
\includegraphics[width=0.5\columnwidth]{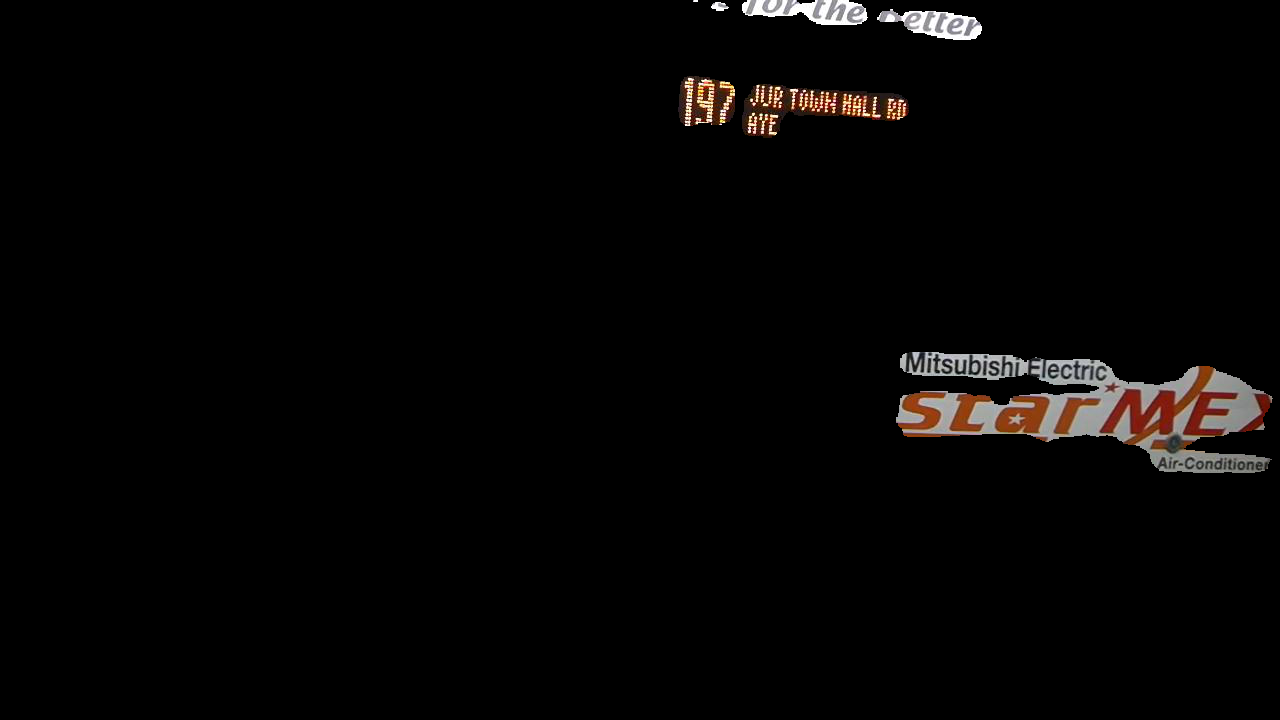}\\
Input image & FCN segmentation
\end{tabular}
\setlength\tabcolsep{6pt} 
\caption{FCN is in general unable to separate individual words.}
\label{fig:FCN_samples}
\end{figure}

 FCN is designed to produce a pixel-level classification, where the label assigned to each pixel comes from a multi-scale analysis of the pixel's neighborhood ({\em scale}  here identifies the effective size of each node's receptive field). The
 ability to utilize both local and extended context allows FCN to produce high quality semantic segmentation. Text patterns, however, are a special category of ``objects", characterized by a specific geometric structure: characters are spaced regularly along a mostly straight line, with words in a line separated by relatively small gaps. With large receptive fields, it is hard for FCN to reliably separate  individual words, resulting in segments that may  contain  a group of text  lines, an individual line, or even an individual word or  character (see Fig. \ref{fig:FCN_samples}).

%

Direct application of object detection algorithms such as R-CNN or YOLO also generally produces poor results. 
R-CNN \cite{girshick2014rich} relies on multiple region proposals, generated by methods such as selective search; this limits both  performance and speed. Its successor, Faster R-CNN \cite{ren2015faster}, replaces  selective search  with a learning-based algorithm for proposal generation. But due to the   large variation in scale and aspect ratio that is typical of word regions,  Faster R-CNN does not produce very good results,  as reported in \cite{tian2016detecting}. It should also be noted that  most object detection algorithms predict axis-aligned rectangular regions, while words may have arbitrary orientation. 

YOLO \cite{redmon2016you} formulates object detection  as a regression problem. This algorithm  is very fast and has shown good results for general object detection, but is  by nature constrained in terms of its ``capacity (the maximum number and density of regions produced in output). YOLO generates two boxes centered at each cell of  a set defined on a regular grid. If more than two regions (e.g. words) are centered within the same cell, they cannot all be detected. The network capacity could be increased by changing the size of the cells or the number of boxes generated per cell. This, however, would result in increased computation and  false positive rate.

Our architecture is a cascade of two stages (see Fig.~\ref{fig:fcn-yolo}): a segmentation stage   (TextSegNet), that detects areas with high likelihood to contain text; and a word detection stage  (WordDetNet), that, from the area cropped out by TextSegNet, resized to a common size, identifies and localizes individual words. The two stages are described in detail in the following.

\begin{figure}[h]
\begin{minipage}[b]{.99\linewidth}
  \centering
  \centerline{\epsfig{figure=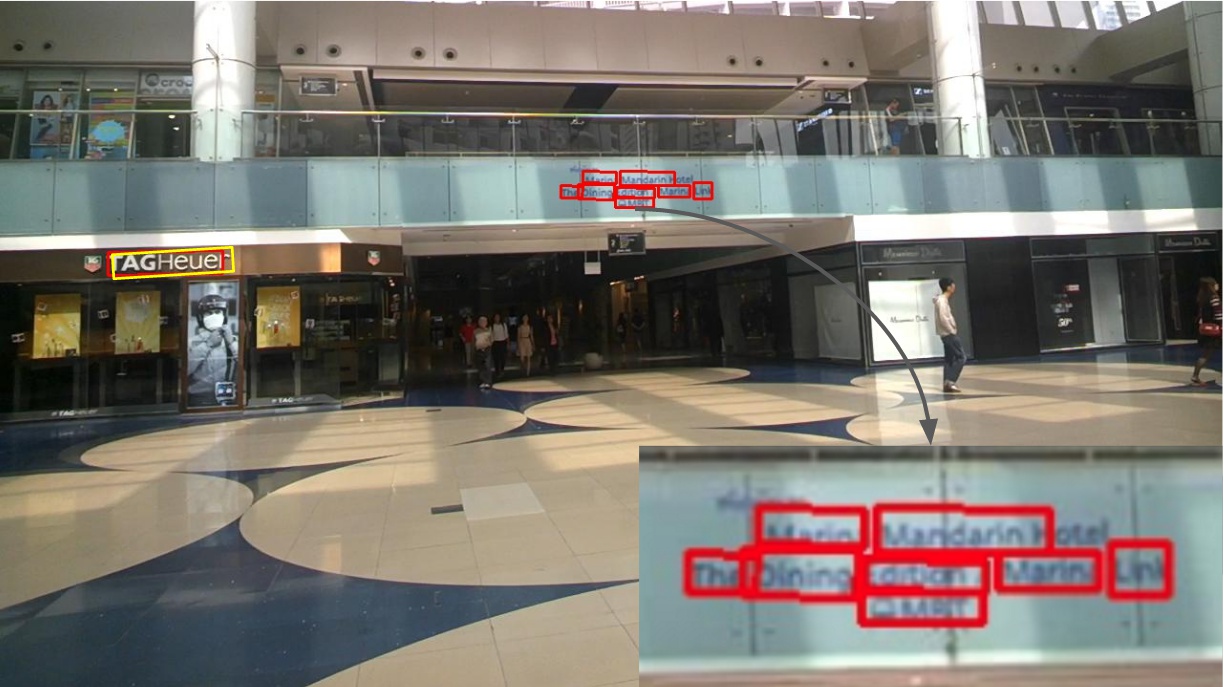,width=8.4cm}}
\end{minipage}
\caption{By cascading  segmentation (TextSegNet) with detection (WordDetNet), our algorithm can detect both large and small words.  If only WordDetNet is used, as trained on whole images,  detection will fail in the case of too small words. Detected words by TextSegNet + WordDetNet are shown by red rectangles, the result of WordDetNet only are shown by yellow rectangles.}
\label{fig:segdet_vs_det}
\end{figure}

\subsection{TextSegNet}
The first stage in our cascade, TextSegNet, is a fully convolutional network	 that takes both local and global context information into consideration to determine the label of each pixel. Text block detection is cast as a semantic segmentation problem with two labels (`text'  and `non-text'). 


\subsubsection{Architecture} 
TextSegNet is based on the  FCN-8s \cite{long2015fully} model, which  is derived from the VGG 16-layer network \cite{simonyan2014very} with the final classifier layer removed, and the fully connected layers converted to convolution layers. We attach an additional final $1\times 1$ convolution layer with channel dimension 2 to obtain prediction scores for `text' and `non-text'. Two skip layers are used to combine finer details (\emph{pool 3} and \emph{pool 4}) with high level semantic information; the output of the main and skip layers are combined and interpolated to the original image resolution using bilinear kernels (the weights of these kernels are also learnt during training). Softmax loss is used for training. The output of the network is a binary map representing likely areas containing text. For more details about the network structure of FCN-8s, the reader is referred to \cite{long2015fully}.

\subsubsection{Training} 
During training and testing, all input images are resized so that their largest side has length of 1000 pixels. Both datasets considered in the experiments have ground-truth word-level labelling. Specifically, individual words are labelled by axis-aligned rectangular bounding boxes in the ICDAR 2013 focused dataset, and by generic quadrilaterals in the ICDAR 2015 incidental dataset. In order to train TextSegNet, a binary mask is first created, where all pixels in the word labelled regions  are marked as `text', while the other pixels are marked as `non-text'.

%

\subsection{WordDetNet}\label{sec:WordDetNet}

Each individual region identified by TextSegNet is first resized to a fixed square shape. More precisely, a square box tightly bounding and co-centered with the region is  computed; the square box is then reshaped uniformly to a fixed  size. In this way, the aspect ratio of the text image is not changed. The only exception is for segments that are very close to the image edges, where a co-centered square bounding box would extend outside the image area. In this case, a rectangular bounding box is considered, which is then re-sized as a square (see eg. Fig.~\ref{fig:fcn-yolo}).
Reshaping text image segments to uniform size provides some degree of scale invariance, except for segments containing one or a few  very long text lines, in which case the reshaped characters may have small size.  The square reshaped images are then fed to a  network (WordDetNet) to predict the locations of individual words. 

Inspired by the YOLO architecture~\cite{redmon2016you}, WordDetNet is tasked with identifying individual words. It  defines a   $N\times N$ grid on the image, where each cell in the grid predicts $B$ candidate oriented rectangular regions (boxes), all centered within the cell. A box can be parameterized in terms of the position $(x,y)$ of its center relative to the bounds of the grid cell, its width and height $(w,h)$ relative to the size of image, and its orientation angle $\theta$, which is normalized to a value between 0 and 1. In addition, the network produces a value ($C$)  that represents the confidence that this box actually contains a word. 

\subsubsection{Loss Function}
The network is trained to minimize a multi-part squared loss function $L({\bf P})$. ${\bf P}$ represents the set of all $N^2\cdot B$ box parameter vectors  ${\bf p}_i^j=( x_i^j, y_i^j, w_i^j, h_i^j, \theta_i^j, C_i^j)$, where $i$ identifies the cell and $j$ identifies the box. $L({\bf P})$ is defined as follows:
\begin{equation}\label{eq:loss}
\begin{split}
L({\bf P}) =  \sum_{i=1}^{N\times N}\delta_i^{obj}  \sum_{j=1}^B \left[ \delta_i^j L^{obj}({\bf p}_i^j) + (1-\delta_i^j)(C_i^j)^2 \right] \\
+\lambda_{noobj}\sum_{i=1}^{N\times N}(1-\delta_i^{obj})\sum_{j=1}^B  (C_i^j)^2
\end{split}
\end{equation}
%
$$
L^{obj}({\bf p}_i^j) =  \left(1-C_i^j\right)^2 + \lambda_{ang}\left(\hat\theta_i-{\theta}_i^j\right)^2
+\lambda_{coord}\,\cdot
$$
$$\left[ (\hat x_i-{x}_i^j)^2+(\hat y_i-{y}_i^j)^2+(\sqrt{\hat w_i}-\sqrt{{w}_i^j})^2+(\sqrt{\hat h_i}-\sqrt{{h}_i^j})^2\right] 
$$
In the equation above, $\delta_i^{obj}$ is equal to 1 if the training image contains a word (represented by a oriented rectangle) centered at the $i$-th cell, 0 otherwise. $\delta_i^j$ is 1 only for the $j$-th  predicted box with the largest Intersection-over-Union (IoU) with the word centered at the $i$-th cell (this is the {\em responsible} box~\cite{redmon2016you}). The hatted notation represents ``ground truth" values for the oriented rectangular word regions. In practice, for cells with no words centered on them, only the second term of the loss expression is activated, which penalizes predicted box confidence values larger than 0. Otherwise, only one responsible box is considered, with a penalty that takes into account both the box's  confidence (which should be close to 1) and its localization accuracy; the non-responsible boxes are given a penalty for large confidence values. Note that Eq.~ (\ref{eq:loss}) is equivalent to the loss function for the original YOLO network, except that (i) it contains an additional term for the rectangle orientation ($\theta$), and (ii) there is no term assessing the posterior distribution of class assignment, as only one class (text) is considered here. We set the weights as follows: $\lambda_{ang}=10$, $\lambda_{coord}=5$ and $\lambda_{noobj}=0.1$ (the smaller value for $\lambda_{noobj}$ is justified by the fact that only few cells are expected  to be the center of  words in the text region). 

Predicted boxes with confidence value $C_i^j$ less than 0.5 are discarded. Non-maximum suppression is used to remove overlapping detection with IoU less than 0.3. Note that general object detection algorithms  use a larger threshold on the IoU. However, since words are normally placed along a line, a smaller overlap is expected in this case. 

\subsubsection{Architecture}
\begin{figure}[h]
\centering
\includegraphics[width=1\columnwidth]{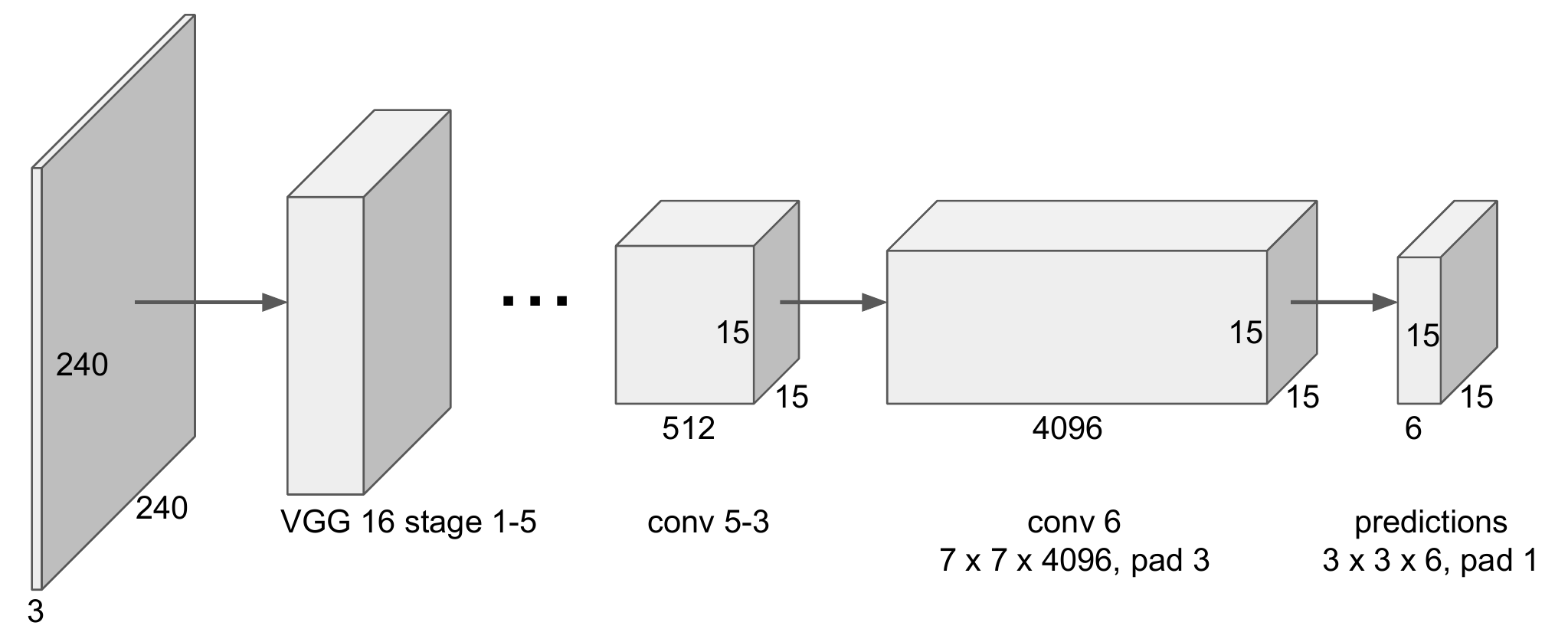}
\caption{The structure of WordDetNet, shown here for $N$=15 and $B$=1. Our experiments show the above configuration achieves the best result.}
\label{fig:detnet}
\end{figure}

WordDetNet (see Fig. \ref{fig:detnet}) is built  on the VGG 16-layer architecture~\cite{simonyan2014very}, with the front layers initialized with the weights learned for TextSegNet (which also based on VGG 16-layer network). The original VGG 16-layer network is decapitated after  \emph{conv 5-3}, and  two additional convolutional layers (\emph{conv 6} and \emph{prediction})  are added, along with a  ReLU and dropout layer after \emph{conv 6}. {\em conv 6} has 4096 filters with kernels size of $7\times 7$, while {\em prediction } has $B\cdot 6$ filters with size of $3\times 3$, resulting in $B$ sets of box parameters ($\{{\bf p}_i^j\}$). Both \emph{conv 6} and \emph{prediction} are properly padded to maintain the size of the feature map. Note that  that there are four $2\times 2$ max pooling layers before  \emph{conv 5-3}. This means that, in order for the channels in output of \emph{conv 5-3} to have size of $N\times N$ (corresponding to the cell grid described earlier), the input image must have size of $16\cdot N \times 16\cdot N$. For example, an input $240\times 240$ color image will result in a grid of $15\times 15$ cells. 
Note that the original YOLO network contains multiple fully connected layers, which are replaced by convolutional layers  in WordDetNet. This results in a smaller number of parameters, allowing for easier training.

\subsubsection{Training} \label{sec:WT}
WordDetNet operates on square text block regions. During training, text blocks are generated by the following algorithm, which mimics the expected output of a typical segmenter, where nearby words are likely to be grouped within the same segment. A graph is formed on the ground truth word-level rectangular bounding boxes, where two such bounding boxes are linked in the graph if their minimum distance is smaller than a threshold (set to be equal to the sum of the heights of the two rectangles). Then, the connected components of this graph are found.  For each connected component, the tightest axis-aligned  bounding rectangle  is computed. Then, as explained in Sec.~\ref{sec:WordDetNet}, this rectangle expanded to a square and resized, before being passed on to WordDetNet for training. 

\begin{figure*}[t]
\setlength\tabcolsep{1.5pt} 
\begin{tabular}{ccc}
\includegraphics[width=0.67\columnwidth]{figures/FCN_YOLO/img_34.jpg}&
\includegraphics[width=0.67\columnwidth]{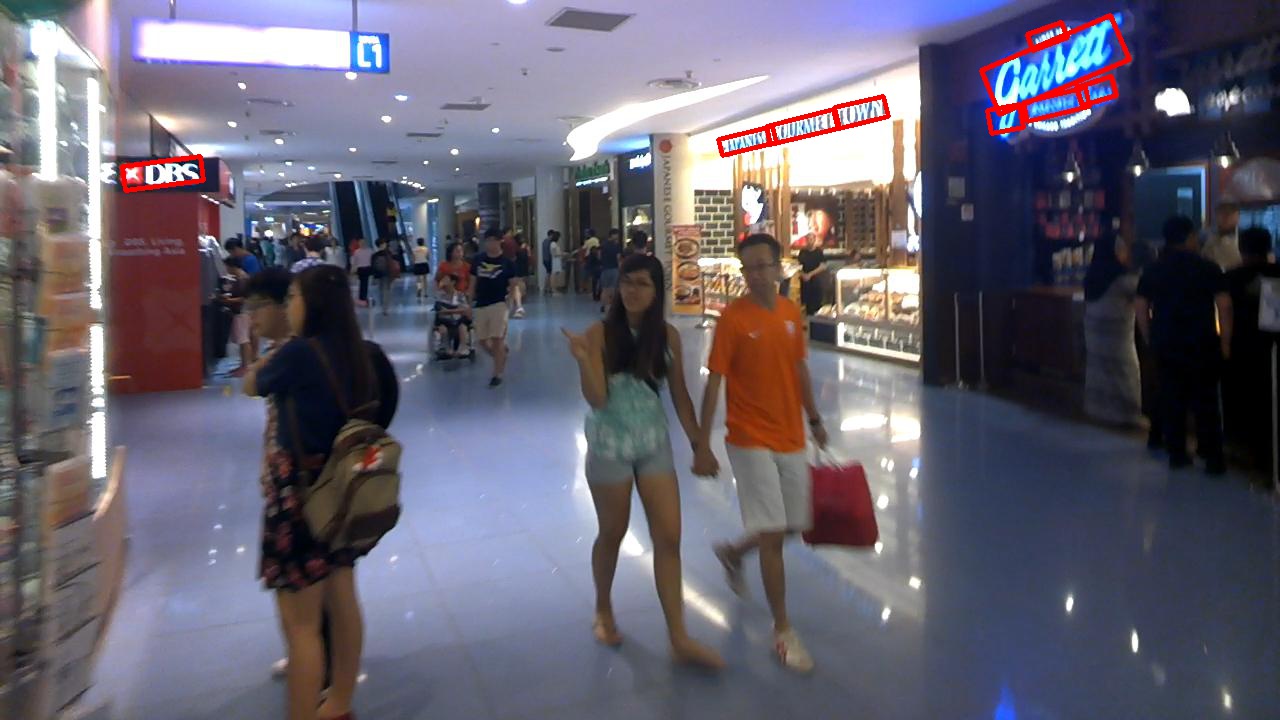}&
\includegraphics[width=0.67\columnwidth]{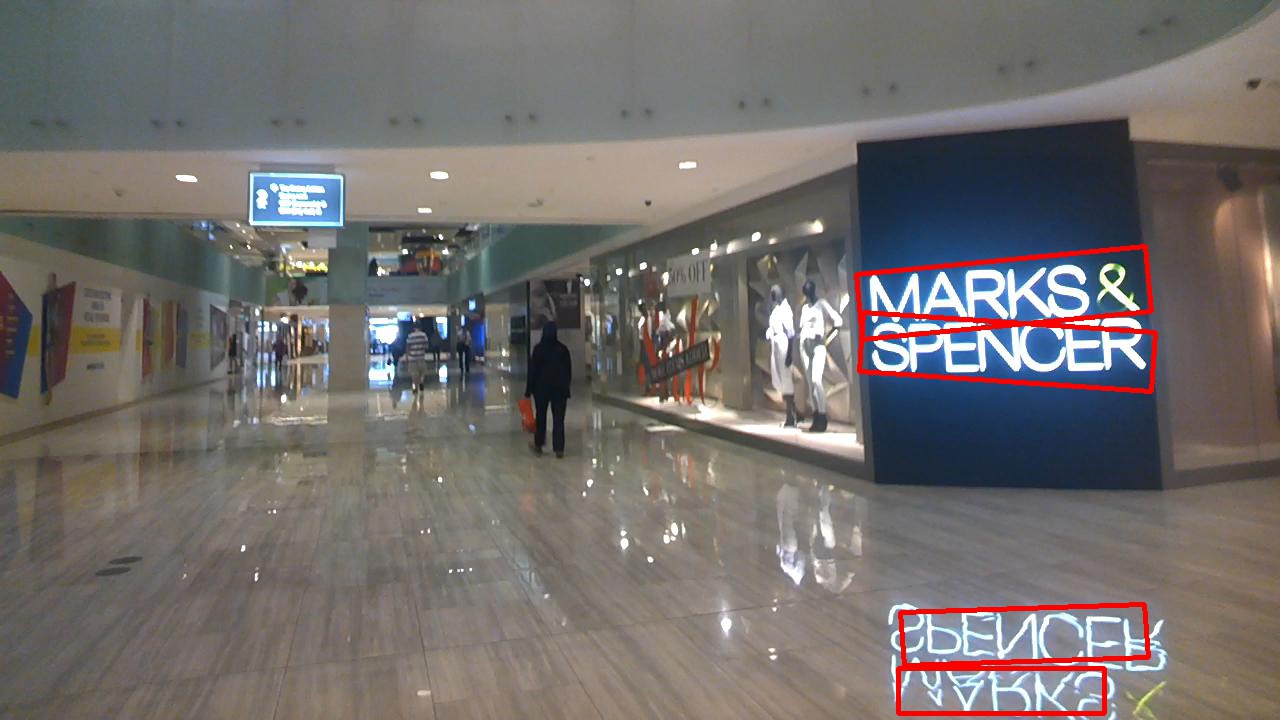}\\
\includegraphics[width=0.67\columnwidth]{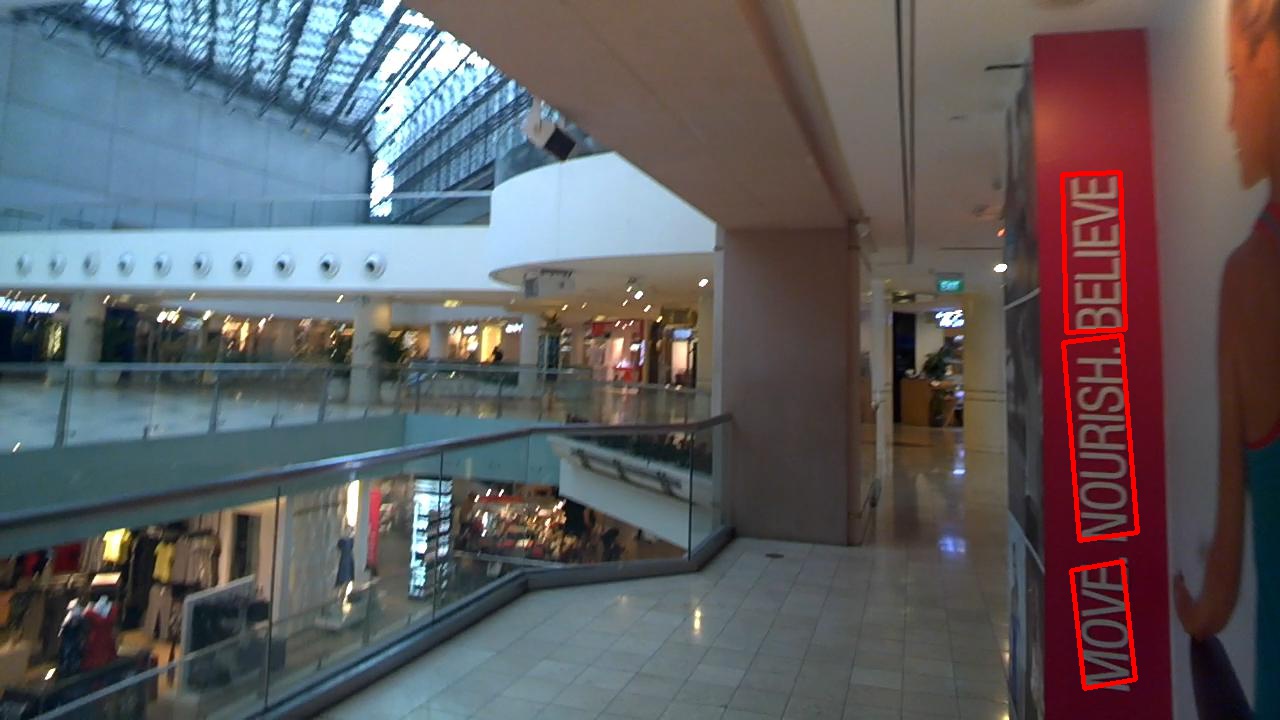}&
\includegraphics[width=0.67\columnwidth]{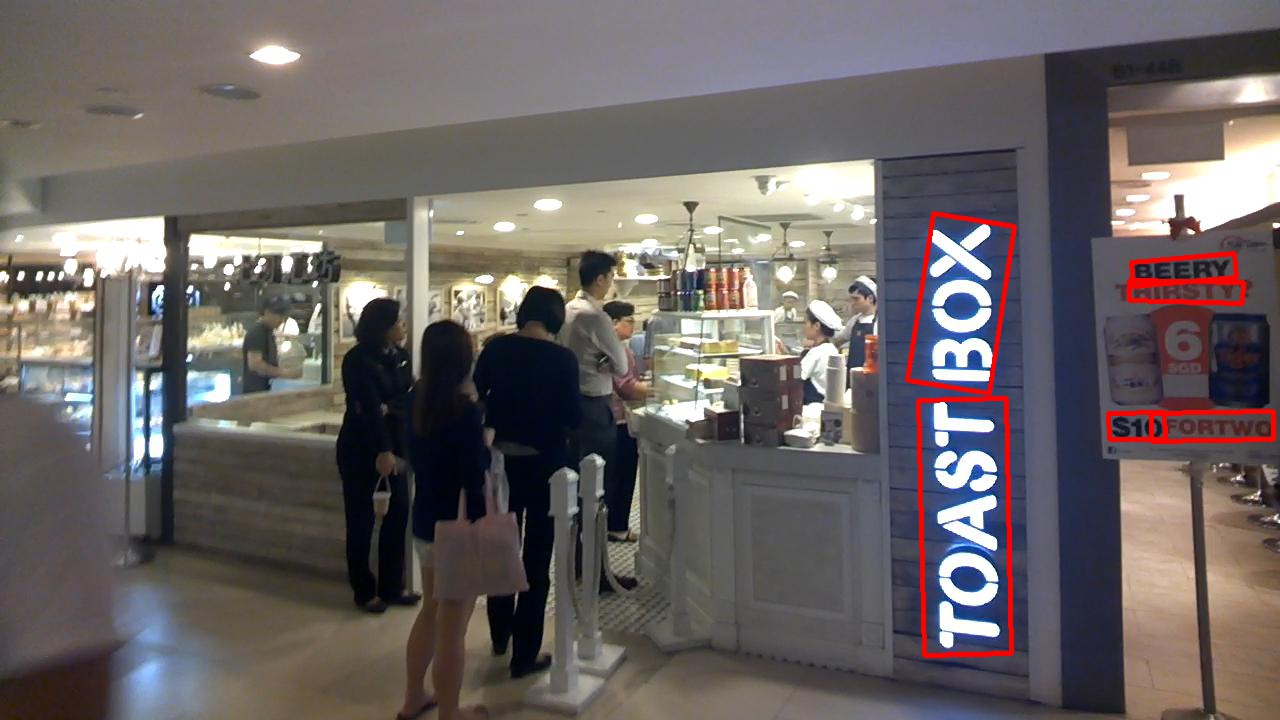}&
\includegraphics[width=0.67\columnwidth]{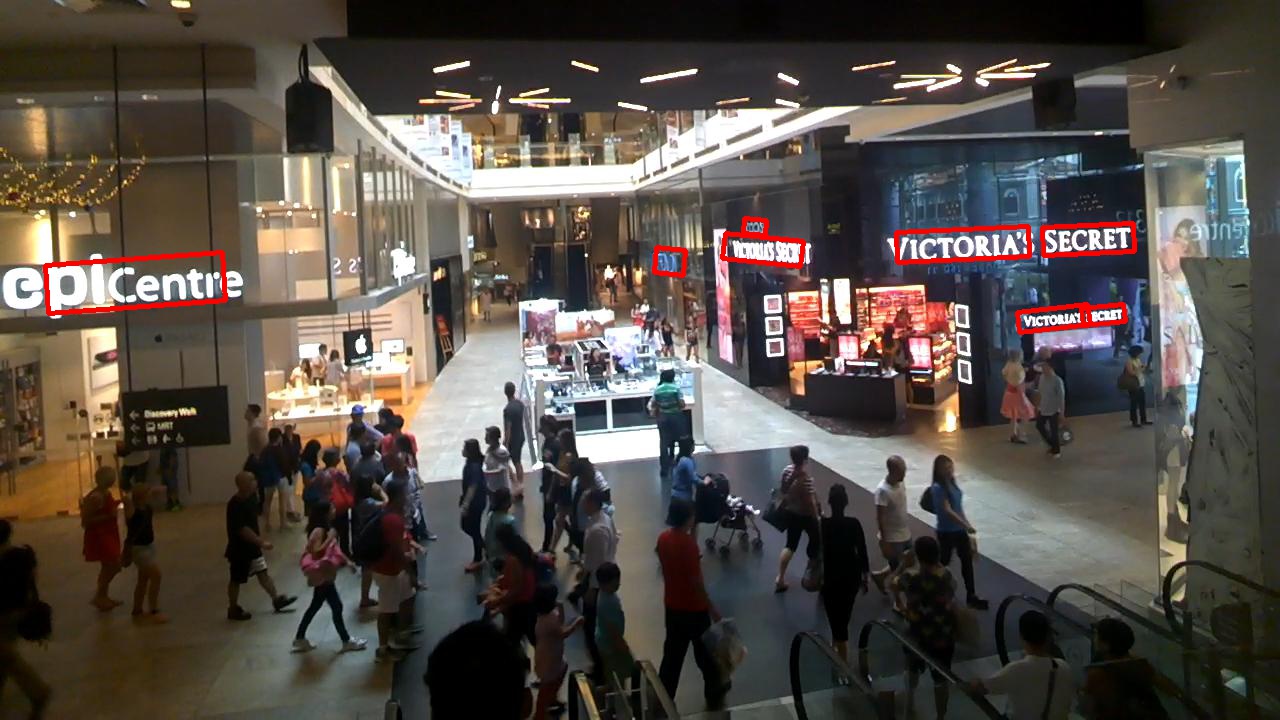}\\
\end{tabular}
\caption{Examples of word-level text spotting on the  ICDAR 2015 Incidental dataset. Detected word regions are shown by red boxes.}
\label{fig:good_results}
\end{figure*}

\section{Experiments}
\subsection{Implementation Details}
Our training data come from three sources:  the training portion of the ICDAR 2013 focused dataset (229 images) and of the ICDAR 2015 incidental dataset (1000 images), as well as the large scale synthetic dataset (~8 million images) described in \cite{gupta2016synthetic}. We augment the images from the ICDAR datasets by means of random rotations, translations, and color adjustment, and add a subset of 20K  images randomly selected   from the synthetic dataset \cite{gupta2016synthetic}. Note that we use only a small portion of the synthetic dataset in order to maintain a balance between natural and synthetic images. Overall, our training dataset contains about 35K images. We found that the  addition of  synthetic images has a moderate effect on performance (F-score increase by 1\% in the ICDAR incidental dataset only). 


The training dataset for WordDetNet contains 40K text blocks which are automatically mined using the algorithm mentioned earlier in Sec.~\ref{sec:WT}.  Weight sharing between TextSegNet and WordDetNet enables good performance in spite of relatively small training data size.

Our system is designed with Caffe \cite{jia2014caffe} on Python, and is implemented on a workstation (3.3Ghz 6-score CPU, 32G RAM, GTX Titan X GPU and Ubuntu 14.04). As mentioned earlier, the images given in input to  TextSegNet are resized to 1000 pixels in their longer side. We use the same training strategy as in \cite{long2015fully}, with batch size of 1, learning rate of $10^{-9}$, momentum of 0.99, and weight decay of 0.0005. Training TextSegNet takes 20 hours for 100K iterations.   WordDetNet takes text blocks resized to  $240\times 240$ pixels, resulting in a $15\times 15$ grid. Training parameters are:  batch size of 16, learning rate of $10^{-5}$, momentum of 0.9 and weight decay of 0.0005. At  50K iterations, training of WordDetNet takes 10 hours.

At deployment, one image is processed by TextSegNet in about  250 ms, then each text block is processed  by WordDetNet in about 50 ms. End-to-end processing takes about 450 ms per image on average.

\subsection{ICDAR 2015 Incidental Dataset}
\subsubsection{Dataset Description}
The ICDAR 2015 incidental dataset \cite{icdarincidental} contains 1000 training images and 500 images used for testing. These images were taken by wearable cameras, without intentional focus on text regions. They are characterized by large variance in text size and orientation; some amount of blur is often visible. This dataset thus represents a much more challenging  benchmark than the older ICDAR 2013 Focused dataset, which is described later in Sec.~\ref{sec:2013}. A quadrilateral bounding box is defined for each word in the dataset; however, only the bounding boxes for the training portion are made available to the public. Detection results are evaluated by measuring the Intersection-over-Union (IoU) between a predicted box and the closest ground truth quadrilateral; if the IoU is larger than 0.5, the predicted box is deemed  a true positive. A score is produced in terms of precision (ratio of true positives count and all detections count) and recall (ratio of true positives count and all ground truth labels count), as well as of their harmonic mean (F-score).  Note that some unreadable words are marked as ``do not care"; they are still counted as ground truth labels when computing precision, but not when computing recall. Our algorithm is only trained on the words not labelled as ``do not care".
\begin{table}[h]
\caption{Results on the ICDAR 2015 Incidental dataset }
\label{tab:incidental}
\begin{tabular}{c||c|c|c}
Method & Precision (\%) & Recall (\%) & F-score (\%)\\ \hline \hline
HUST \cite{karatzas2015icdar} & 44 & 38 & 41 \\ \hline
AJON \cite{karatzas2015icdar} & 47 & 47 & 47 \\ \hline
NJU-Text \cite{karatzas2015icdar} & 70 & 36 & 47 \\ \hline
StradVision \cite{karatzas2015icdar} & 53 & 46 & 50 \\ \hline
Zhang \cite{zhang2016multi} & 71 & 43 & 54 \\ \hline
Google Vision API~\footnotemark & 68 & 53 & 59 \\ \hline
CTPN \cite{tian2016detecting} & 74 & 52 & 61 \\ \hline
Megvii-image++ \cite{yao2016scene} & 72 & 58 & 64 \\ \hline \hline
{\bf Proposed (Seg+Det)} & {\bf 79} & {\bf 65} & {\bf 71} \\ \hline
Proposed (Det only) & 61 & 40 & 48 \\ \hline
\end{tabular} \label{tab:Incidental}
\end{table}

\subsubsection{Results}
Tab.~\ref{tab:Incidental} shows  results of our method as compared with other state of the art published algorithms~\cite{zhang2016multi,tian2016detecting,yao2016scene}, as well as with Google Vision API~\footnotemark[\value{footnote}], which produces word-level detection and recognition. 
Our cascaded  (segmentation + detection) network outperforms the previous best results by a large margin (7\% higher F-score than its closest competitor). In order to highlight the importance of the prior segmentation step, we also show results  using only WordDetNet as applied on the whole image, rather than on the segments detected by TextSegNet. Performance decreases substantially in this case, showing the importance of a prior segmentation step. Some detection examples in challenging images are shown in Fig.~\ref{fig:good_results}.

\footnotetext{\url{https://cloud.google.com/vision/}}

\begin{figure}[t!!]
\centering
\includegraphics[width=0.45\textwidth]{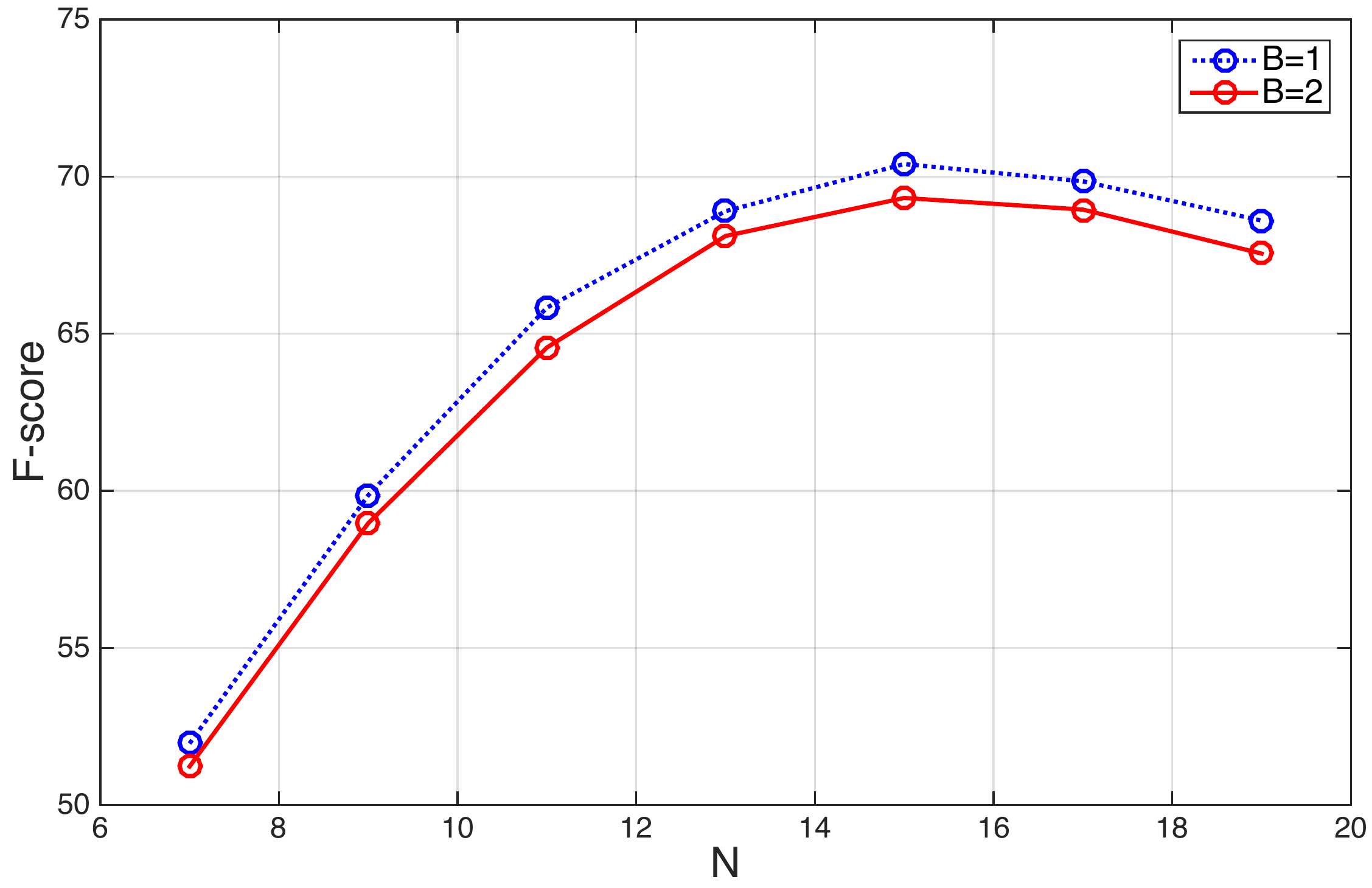}
\caption{The F-score of ours system on the ICDAR  2015 incidental  dataset as a function of the parameters $N$  (which determines the number of cells in grid considered by WordDetNet) and $B$ (the number of boxes generated per cell).}
\label{fig:S_C}
\end{figure}

The network capacities (defined as the maximum density of candidate boxes generated) can be changed  by varying the number of cells in the $N\times N$ grid defined in WordDetNet, or the number $B$ of boxes generated in each cell. Fig.~\ref{fig:S_C} plots the F-score resulting from varying $N$ between 7 and 19  (only odd values ~\cite{redmon2016you}) and with $B$ equal to 1 and 2. This data shows that generating more than one candidate box per cell doesn't seem to provide an advantage, and that the optimal number of cells is $15\times 15$. Note that with fewer cells, the burden is on the network to correctly localize the box within a larger cell. With more (hence smaller) cells, correct  localization is easier, but the risk of false positives increases.

\begin{figure}[h]
\begin{minipage}[b]{.49\linewidth}
  \centering
  \centerline{\epsfig{figure=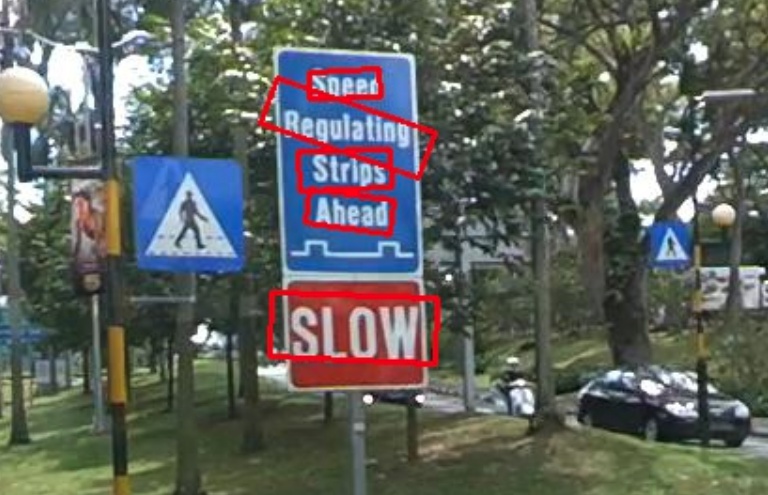,width=4.4cm}}
\end{minipage}
\hfill
\begin{minipage}[b]{.49\linewidth}
  \centering
  \centerline{\epsfig{figure=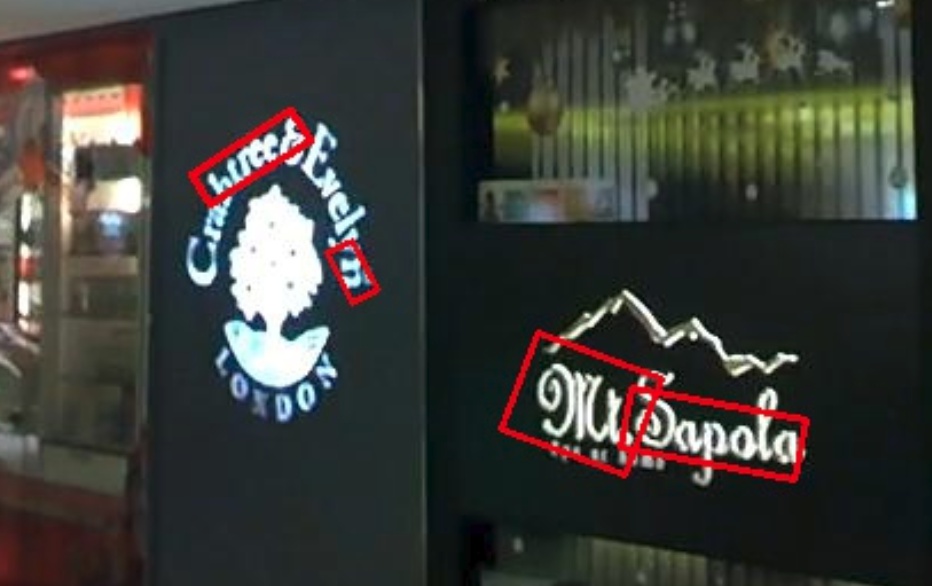,width=4.4cm}}
\end{minipage}
\caption{Examples of failure cases: incorrect box orientation or size, missing words in curved text.}
\label{fig:bad_results}
\end{figure}

In spite of the good quantitative results, we noticed that sometime the box orientation and/or size as estimated by WordDetNet is somewhat inaccurate (see examples in Fig.~\ref{fig:bad_results}). Incorrect orientation or size of an estimated word region may reduce the Intersection-over-Union with the corresponding ground truth region, thus affecting the resulting recall rate.

\begin{table}[h]
\caption{Results on the ICDAR 2013 Focused dataset}
\label{tab:focused}
\begin{tabular}{c||c|c|c}
Method & Precision (\%) & Recall (\%) & F-score (\%)\\ \hline \hline
Neumann \textit{et al.} \cite{neumann2013combining} & 85 & 68 & 75 \\ \hline
Yin \textit{et al.} \cite{yin2014robust} & 86 & 68 & 76 \\ \hline
FASText \cite{busta2015fastext} & 84 & 69 & 77 \\ \hline
Huang \textit{et al.} \cite{huang2014robust} & 88 & 71 & 78 \\ \hline
Zhang \textit{et al.} \cite{zhang2015symmetry} & 88 & 74 & 80 \\ \hline
TextFlow \cite{tian2015text} & 85 & 76 & 80 \\ \hline
He \textit{et al.} \cite{he2016text} & 93 & 73 & 82 \\ \hline
Qin \textit{et al.} \cite{qin2016wacv} & 88 & 77 & 82 \\ \hline
Zhang \textit{et al.} \cite{zhang2016multi} & 88 & 78 & 83 \\ \hline
Gupta \textit{et al.} \cite{gupta2016synthetic} & 92 & 76 & 83 \\ \hline
Yao \textit{et al.} \cite{yao2016scene} & 88 & 80 & 84 \\ \hline
TextBoxes \cite{liao2016textboxes} & 88 & {\bf 83} & 85 \\ \hline
Zhu \textit{et al.} \cite{zhu2016text} & {\bf 93} & 81 & 87 \\ \hline
CTPN \cite{tian2016detecting} & {\bf 93} & {\bf 83} & {\bf 88} \\ \hline \hline
{\bf Proposed} & 90 & {\bf 83} & 86 \\ \hline
\end{tabular}
\end{table}

\subsection{ICDAR 2013 Focused Dataset}\label{sec:2013}

\subsubsection{Dataset Description}

The ICDAR 2013 focused dataset \cite{icdarfocused} contains  images ``explicitly focused around the text content of interest". 229 images are used for training and 233 for testing. In these images,   text is seen at good resolution and at approximately  horizontal orientation. 
 Each word is labelled with an axis-aligned rectangle. The evaluation protocol is described in  \cite{karatzas2013icdar,wolf2006object}.

\subsubsection{Results}
\begin{figure}[h]
\centering
\includegraphics[width=0.45\textwidth]{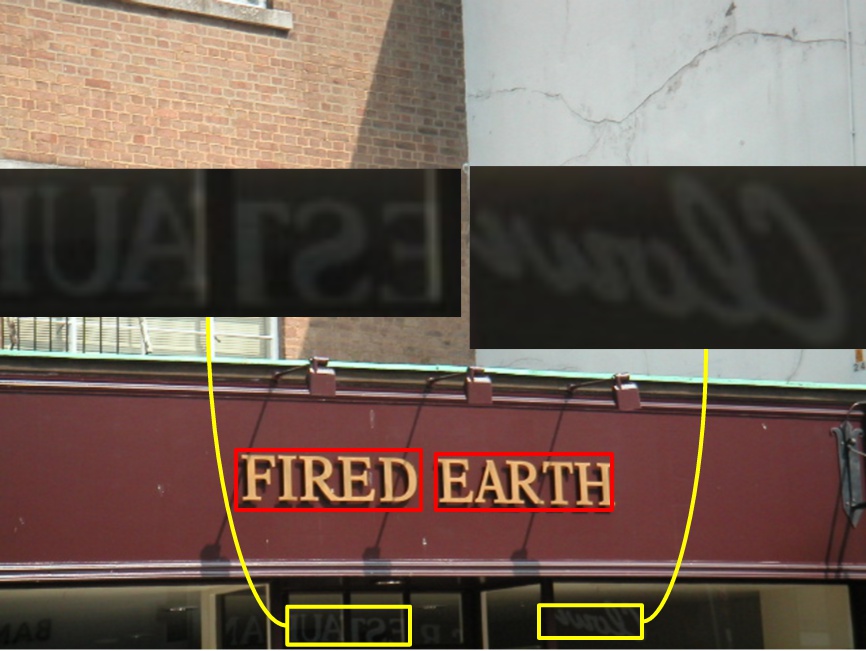}
\caption{Our method  finds words in  very challenging situations (yellow boxes), even when these words are not labelled in the ICDAR 2013 focused dataset.}
\label{fig:unlabelled}
\end{figure}

Tab.~\ref{tab:focused} shows comparative results against other  published algorithms. Our system has  F-score of 86\%, ranking among the top performers. The best current result is achieved by CTPN \cite{tian2016detecting}, with  F-score of 88\%. Note, however, that CTPN cannot identify individual words, and does not work for text with arbitrary orientation. In the more challenging ICDAR 2015 incidental dataset, our system outperforms CTPN by 10\%. Looking closer at the data, one may notice that our algorithm produces the same recall value (83\%) as the best performing systems, but lags behind in precision (90\%, vs. 93\% as obtained with CTPN \cite{tian2016detecting} and Zhu \textit{et al.} \cite{zhu2016text}). Part of the reason is that  our method is able to find existing text that doesn't appear in the  ground-truth labelling (see e.g. Fig.~\ref{fig:unlabelled}), resulting in an (incorrect) penalty in terms of precision.

\section{Conclusion}
We have presented a new approach for word-level text detection and localization. Our algorithm identifies individual words and draws bounding boxes in the shape of oriented rectangles. The system is formed by the cascaded of a segmentation network (TextSegNet) and a detection network (WordDetNet), where the latter operates on regions segmented out by the former, resized to a common size. By combining segmentation and detection, we leverage on the strengths of each network. TextSegNet (which is based on the fully convolutional architecture of~\cite{long2015fully}) can very robustly detect the presence of text in the image, but is unable to identify individual words. WordDetNet (inspired by the YOLO architecture~\cite{redmon2016you}) can effectively detect and localize individual words from a resized regions identified by  TextSegNet. Unlike most existing algorithms,  our system does not require a post-processing step to enforce alignment of the detected words.

We show experimentally that the first step (segmentation and resizing) is critical for the second step to be effective. On the challenging ICDAR 2015 incidental dataset, our system achieves top results among the published algorithms, outperforming the closest one by 7\% in terms of F-score. In the more benign  ICDAR 2013 focused dataset, our system produces an F-score that is only 2\% less than the top performing algorithm (CTPN \cite{tian2016detecting}).


%
%
%



%
\bibliographystyle{IEEEtran}
\bibliography{Siyang}


\end{document}